\documentclass[11pt]{article}

\usepackage[preprint]{acl}

\usepackage{times}
\usepackage{latexsym}
\usepackage[T1]{fontenc}
\usepackage[utf8]{inputenc}
\usepackage{microtype}
\usepackage{inconsolata}
\usepackage{graphicx}
\usepackage{booktabs}
\usepackage{multirow}
\usepackage{amsmath}
\usepackage{float}
\usepackage{tipa}
\usepackage{adjustbox}
\usepackage{tikz}
\usetikzlibrary{positioning,fit,arrows.meta,calc}
\usepackage{pgfplots}
\pgfplotsset{compat=1.17}

\title{WaxalNet: Benchmarking and Adapting Speech Recognition\\ for Spontaneous Speech in 19 African Languages}

\author{
\small
  \begin{tabular}{c}
    
    Victor Tolulope Olufemi$^{1,2}$, Oreoluwa Babatunde$^{2}$, Ramsey Njema$^{1}$, Bolarinwa Gbotemi$^{2}$, Wanchi Lucia Yen$^{1}$, \\
    John Uzodinma$^{1}$, Sunday Ajayi$^{1}$, Oluwademilade Williams$^{2}$, Kausar Moshood$^{2}$, Innocent Elendu Anyaele$^{1}$, \\
    Akebert Arefaine$^{1}$, Candace Hunzwi$^{1}$, Wongel Dawit Daniel$^{1}$, Emmilly Namuganga$^{1}$, Cleophas Kadima$^{1}$, \\
    Athanase Bahizire$^{1}$, Onitsiky Ranaivoson$^{1}$, Emmanuel Aaron$^{1}$, Nicholaus Ladislaus$^{1}$, Idris Muhammed$^{1}$, \\
    Jonathan Enoch Simenya$^{1}$, Martin Koome$^{1}$, Matewos Tegete Endaylalu$^{1}$, Peter Ifeoluwa Adeyemo$^{1}$, \\
    Hondi Prisca Birindwa$^{1}$, Ukachi Agnes Eze-Mbey$^{1}$, Yacoba Oduro-Yeboah$^{1}$, \\
    Pericles Adjovi$^{1}$, Mikel K.\ Ngueajio$^{1}$, Toluwani Aremu$^{3}$, Prasenjit Mitra$^{1}$
  \end{tabular}
  \\[0.4em]
  \small
  $^{1}$CMU Africa \quad
  $^{2}$LyngualLabs \quad
  $^{3}$MBZUAI
}

\begin{document}
\maketitle

\begin{abstract}
Multilingual speech foundation models claim broad language coverage, but how well they handle spontaneous African speech, and how best to adapt them, remains unclear. We benchmark zero-shot and fine-tuned speech recognition on the WAXAL corpus of spontaneous speech in 19 African languages, covering Whisper, MMS, Omnilingual ASR, and SeamlessM4T-v2. Fine-tuning lowers macro word error rate (WER) from 52.3 for the best zero-shot model covering all 19 languages (Omnilingual-7B) to 36.2 for a 300M-parameter MMS model and 31.5 for Omnilingual with an LLM decoder, although zero-shot models are already competitive on a few languages. Using Omnilingual's shared encoder, we compare CTC and LLM decoders with encoder pretraining held fixed. Between MMS and Whisper, the better model depends on the language family, but with a shared encoder the LLM decoder has lower WER on 18 of 19 languages, so the family pattern more likely comes from pretraining than from the decoder. The LLM decoder is, however, about five times larger and much slower on CPU. Pooling languages into one model, by family or across all 19, costs only 0.6 to 1.0 WER on average relative to per-language models, a modest price for serving all languages with one model. Fine-tuned models remain usable on read speech, and for held-out languages neither training on the other languages nor in-context examples replace in-language fine-tuning. A native-speaker audit across all 19 languages characterizes failure modes such as repetition loops. We release the code and a suite of over 100 fine-tuned models.\footnote{Models: \url{https://huggingface.co/collections/waxal-benchmarking/waxalnet-6ac0f413f74b466ebbf09c96}}
\end{abstract}

\section{Introduction}

Systems such as Whisper~\citep{radford2023robust}, Massively Multilingual Speech (MMS)~\citep{pratap2024scaling}, and Omnilingual ASR~\citep{omnilingual2025} now list hundreds to thousands of supported languages. For many African languages, however, recognition quality and deployability still lag far behind~\citep{olatunji2023afrispeech, conneau2023fleurs}. Two gaps stand out. The first is accuracy: multilingual models often struggle with spontaneous speech, code-switching, tone, and dialectal variation, and sometimes produce hallucination loops or do not support the language at all~\citep{koudounas2025hallucinationbenchmarkspeechfoundation, baranski, atwany-etal-2025-lost}. The second is efficiency: the largest models are impractical on the low-cost devices common where much African speech data is collected.

Fine-tuning is the standard response, and it is known to help low-resource ASR~\citep{imam-etal-2026-full, liu_exploration_2024, emezue25_interspeech}. Several practical questions remain open, however, for anyone building recognition for many languages at once. \emph{(Q1)} Which decoding architecture works best, and does the answer depend on the language? \emph{(Q2)} Should each language get its own model, or can one multilingual model serve them all? \emph{(Q3)} Does a fine-tuned model hold up on a different speech domain, and \emph{(Q4)} can it handle a language it was not trained on?

We study these questions on WAXAL~\citep{Diacketal2026}, a corpus of spontaneous, image-prompted speech in 19 typologically diverse African languages. We evaluate five zero-shot baselines and five fine-tuned models, and use Omnilingual ASR, whose CTC and LLM decoders share one pretrained encoder, to run comparisons in which encoder pretraining is held fixed. All models are scored on the same test utterances with one normalization, and differences are tested with a paired bootstrap. Our contributions are:

\begin{itemize}
\itemsep0.05em
\item A benchmark of zero-shot and fine-tuned ASR on spontaneous speech in 19 African languages, several with little prior ASR evaluation, with all models scored on one test set under one normalization and compared with paired significance tests, and a released suite of over 100 fine-tuned models (\S\ref{sec:bench}).
\item A controlled comparison of decoders: a language-family pattern that appears between MMS (CTC) and Whisper (autoregressive) does not appear when CTC and LLM decoders share one pretrained encoder, which suggests that cross-model comparisons can attribute pretraining differences to architecture (\S\ref{sec:arch}).
\item Practical guidance for building recognition for many languages: the accuracy and CPU cost of each model type, a matched-budget comparison of per-language, per-family, and all-language training, and how fine-tuned models behave on read speech and on languages they were not trained on, including in-context adaptation (\S\ref{sec:arch}--\S\ref{sec:gen}).
\item A native-speaker audit across all 19 languages that characterizes failure modes and script-related evaluation issues (\S\ref{sec:audit}).
\end{itemize}

\definecolor{cBantu}{RGB}{78,121,167}
\definecolor{cKwa}{RGB}{242,142,43}
\definecolor{cGur}{RGB}{225,87,89}
\definecolor{cCush}{RGB}{89,161,79}
\definecolor{cSem}{RGB}{176,122,161}
\definecolor{cIso}{RGB}{150,150,150}
\begin{figure*}[t]
\centering
\begin{minipage}[c]{0.36\textwidth}
\centering
\begin{adjustbox}{max width=\linewidth}
\begin{tikzpicture}[
  font=\footnotesize,
  box/.style={draw, rounded corners=2pt, align=center, inner sep=3pt, minimum height=0.7cm},
  arr/.style={-{Latex[length=1.6mm]}, thick}]
  \node[box] (aud) {Speech};
  \node[box, right=0.35cm of aud, minimum height=1.9cm, fill=gray!12] (enc)
       {Shared\\ \texttt{wav2vec2}\\ encoder};
  \node[box, right=0.4cm of enc.east, yshift=0.5cm, anchor=west, fill=cBantu!15, minimum width=1.95cm] (ctc) {CTC head};
  \node[box, right=0.4cm of enc.east, yshift=-0.5cm, anchor=west, fill=cKwa!15, minimum width=1.95cm] (llm) {LLM decoder};
  \node[box, anchor=west] (txt) at ($(ctc.east)!0.5!(llm.east) + (0.55,0)$) {Text};
  \draw[arr] (aud) -- (enc);
  \draw[arr] (enc.east |- ctc) -- (ctc);
  \draw[arr] (enc.east |- llm) -- (llm);
  \draw[arr] (ctc.east) -- ++(0.22,0) |- ([yshift=0.12cm]txt.west);
  \draw[arr] (llm.east) -- ++(0.22,0) |- ([yshift=-0.12cm]txt.west);
  \node[below=0.15cm of enc, align=center, text width=3.2cm, font=\scriptsize\itshape]
       {pretrained once;\\ decoder varied (\S\ref{sec:arch})};
\end{tikzpicture}
\end{adjustbox}

\vspace{2pt}{\small (a) Decoder comparison}
\end{minipage}%
\hfill
\begin{minipage}[c]{0.62\textwidth}
\centering
\begin{adjustbox}{max width=\linewidth}
\begin{tikzpicture}[font=\footnotesize]
  \def\xs{0/cBantu,0.28/cBantu,0.56/cBantu,0.84/cBantu,1.12/cBantu,1.40/cBantu,
          1.86/cKwa,2.14/cKwa,2.42/cKwa,
          2.88/cGur,3.16/cGur,
          3.62/cCush,3.90/cCush,
          4.36/cSem,4.64/cSem,
          5.10/cIso,5.38/cIso,5.66/cIso,5.94/cIso}
  \node[anchor=east, align=right] at (-0.3,0) {Per-language\\ \scriptsize(19 models)};
  \foreach \x/\c in \xs {
    \draw[gray!60, rounded corners=1pt] (\x-0.12,-0.16) rectangle (\x+0.12,0.16);
    \fill[\c] (\x,0) circle (0.085);}
  \def\y{-0.8}
  \node[anchor=east, align=right] at (-0.3,\y) {Per-family\\ \scriptsize(5 models)};
  \foreach \a/\b in {0/1.40, 1.86/2.42, 2.88/3.16, 3.62/3.90, 4.36/4.64}
    \draw[gray!60, rounded corners=2pt] (\a-0.14,\y-0.17) rectangle (\b+0.14,\y+0.17);
  \foreach \x/\c in \xs \fill[\c] (\x,\y) circle (0.085);
  \def\y{-1.6}
  \node[anchor=east, align=right] at (-0.3,\y) {All 19\\ \scriptsize(1 model)};
  \draw[gray!60, rounded corners=2pt] (-0.14,\y-0.17) rectangle (6.08,\y+0.17);
  \foreach \x/\c in \xs \fill[\c] (\x,\y) circle (0.085);
  \def\y{-2.4}
  \node[anchor=east, align=right] at (-0.3,\y) {Leave-one-out\\ \scriptsize(7 held-out langs.)};
  \draw[gray!60, rounded corners=2pt] (-0.14,\y-0.17) rectangle (6.08,\y+0.17);
  \foreach \x/\c in \xs {
    \ifdim\x pt=0.84pt
      \draw[\c, line width=0.9pt, fill=white] (\x,\y) circle (0.08);
    \else
      \fill[\c] (\x,\y) circle (0.085);
    \fi}
  \draw[-{Latex[length=1.4mm]}, thick] (0.84,\y-0.45) -- (0.84,\y-0.12);
  \node[anchor=west, font=\scriptsize\itshape] at (0.95,\y-0.45) {held out: train on 18, test on 1};
  \def\y{-3.3}
  \foreach \x/\c/\l in {0/cBantu/Bantu, 1.1/cKwa/Kwa, 1.95/cGur/Gur, 2.75/cCush/Cushitic,
                        3.95/cSem/Semitic, 5.2/cIso/no relative} {
    \fill[\c] (\x,\y) circle (0.085);
    \node[anchor=west, font=\scriptsize] at (\x+0.1,\y) {\l};}
\end{tikzpicture}
\end{adjustbox}

\vspace{2pt}{\small (b) Training regimes over the 19 WAXAL languages}
\end{minipage}
\caption{Study overview. (a) Omnilingual ASR pairs one pretrained encoder with interchangeable CTC and LLM decoders, so the decoder can be compared with encoder pretraining held fixed. (b) Each dot is a WAXAL language, colored by family; boxes show which languages share a model in each training regime. In leave-one-out, the hollow language is excluded from training and used only for evaluation.}
\label{fig:overview}
\end{figure*}

\section{Related Work}

\paragraph{Multilingual ASR.} Whisper~\citep{radford2023robust} is an encoder--decoder model trained on 680k hours of weakly supervised web data covering 99 languages. MMS~\citep{pratap2024scaling} extends CTC recognition to over 1,100 languages via \texttt{wav2vec2} pretraining and language adapters. Omnilingual ASR~\citep{omnilingual2025} covers over 1,600 languages and offers CTC and LLM decoders on a shared encoder, as well as a 7B model that adapts to new languages from in-context examples. SeamlessM4T-v2~\citep{barrault2023seamless} supports speech recognition in about 100 languages. Few of these have been evaluated on spontaneous African speech.

\paragraph{African speech resources.} Common Voice~\citep{ardila2020commonvoice} and FLEURS~\citep{conneau2023fleurs} include African languages but consist of read speech. AfriSpeech-200~\citep{olatunji2023afrispeech} covers African-accented English. Voice of a Continent~\citep{elmadany-etal-2025-voice} and AfriVox~\citep{awobade-etal-2026-afrivox} survey model performance across many African languages. WAXAL~\citep{Diacketal2026} offers spontaneous speech recorded in natural environments, and earlier work showed that compact CTC models fine-tuned on WAXAL subsets are competitive~\citep{ehioasr}.

\paragraph{Adaptation and transfer.} Prior work compares full and parameter-efficient fine-tuning for Whisper~\citep{liu_exploration_2024} and African languages~\citep{imam-etal-2026-full}, and shows that multistage multilingual fine-tuning helps when languages are similar~\citep{10.1145/3813800}. Single multilingual ASR models have long been shown to approach or match monolingual models, with gains for low-resource languages and some loss for high-resource ones~\citep{toshniwal2018multilingual, kannan2019large, pratap2020massively}. Similar capacity trade-offs appear in multilingual text models~\citep{conneau2020unsupervised}, and fine-tuning is known to erode previously learned abilities~\citep{kirkpatrick2017overcoming}.

\section{Data and Setup}
\label{sec:setup}

\paragraph{WAXAL.} We use the ASR subset of WAXAL, which contains 446k transcribed utterances (2,279 hours) in training, validation, and test splits defined by the corpus authors with no speaker shared between splits, released under CC-BY-4.0~\citep{Diacketal2026}. After cleaning (below), we use 362k training, 36k validation, and 40k test utterances. Per-language statistics and families are in Appendix~\ref{app:dataset}.

\paragraph{Cleaning and scoring.} Preliminary evaluation revealed reference artifacts that inflate WER. We remove utterances shorter than 1.5\,s and utterances whose transcript would require more than 4 words per second. This mainly removes truncated or misaligned references. All transcripts are NFC-normalized, lower-cased, and stripped of punctuation, keeping diacritics. Before scoring, we also remove apostrophes and replace hyphens with spaces in references and hypotheses, so that differences in how words are joined (elided forms written with an apostrophe, or compounds written with a hyphen) are not counted as recognition errors. We report WER and character error rate (CER).\footnote{WER can exceed 100 when a model inserts many words, as in hallucination loops~\citep{koudounas2025hallucinationbenchmarkspeechfoundation}.}

\paragraph{Evaluation set and significance.} Omnilingual models cannot transcribe clips longer than 40\,s, so every model is scored on the same test utterances of 1.5 to 40\,s, also excluding the few clips on which any model's inference failed. This removes 212 of 39,566 test utterances, at most 51 in one language (Soga). We compare systems with a paired bootstrap over test utterances (2,000 resamples) and report 95\% intervals for WER differences; a per-language difference is called significant when its interval excludes zero. Each model was trained once, so these intervals reflect test-set sampling, not variation across training runs.

\paragraph{Zero-shot baselines.} We evaluate Whisper-Large-v3 (1.5B), MMS-1B, Omnilingual-1B (CTC decoder), Omnilingual-7B (LLM decoder), and SeamlessM4T-v2 (2.3B) without WAXAL training. Omnilingual's pretraining includes all 19 WAXAL languages~\citep{omnilingual2025}, so for Omnilingual these results reflect the pretrained model rather than a true zero-shot setting. MMS-1B (\texttt{mms-1b-all}) is run with each language's adapter and vocabulary, which exist for 16 of the 19 languages; Whisper-Large-v3 is run with the language token forced and supports 4 languages; SeamlessM4T-v2 supports 5. Because autoregressive decoding of full test sets with the 7B and SeamlessM4T-v2 models is slow, we evaluate them on a fixed random sample of 500 test utterances per language; on models evaluated both ways, sample and full-test macro WER differ by less than 0.3 points (Appendix~\ref{app:sample}).

\paragraph{Fine-tuned models.} We fine-tune three edge-sized models: Whisper-Tiny (38M), Whisper-Small (242M), and MMS-300M (CTC, with the encoder frozen). These use an effective batch size of 32, learning rate $10^{-4}$ with 500 warmup steps, and up to 30 epochs with early stopping. We also fully fine-tune Omnilingual's 300M encoder with either its CTC head or its LLM decoder for 5,000 steps (learning rates $10^{-5}$ and $5\times10^{-5}$). The LLM decoder adds 1.3B parameters, so this model has 1.6B parameters in total. Hyperparameters are fixed across languages. Training details are in Appendix~\ref{app:training}.

\section{Benchmark Results}
\label{sec:bench}

\begin{table*}[t]
\centering\small
\setlength{\tabcolsep}{4pt}
\begin{adjustbox}{max width=\textwidth}
\begin{tabular}{lr|cccc|ccccc}
\toprule
& & \multicolumn{4}{c|}{\textbf{Zero-shot / pretrained}} & \multicolumn{5}{c}{\textbf{Fine-tuned on WAXAL}} \\
\textbf{Language} & \textbf{$n$} & Whisper-L3 & MMS-1B & Omni-1B & Omni-7B$^{\ast}$ & Whisper-T & Whisper-S & MMS-300M & Omni-CTC & Omni-LLM \\
\midrule
Acholi    & 506 & n/a & 51.7 & 75.0 & 60.7 & 59.6 & 43.9 & 41.5 & 37.0 & \textbf{33.1} \\
Akan      & 1,522 & n/a & 57.0 & 59.3 & 47.2 & 38.6 & 33.4 & 33.9 & 32.4 & \textbf{29.2} \\
Amharic   & 3,420 & 117.3 & 50.5 & 43.2 & 34.1 & 39.2 & 29.6 & 32.7 & 34.8 & \textbf{25.0} \\
Dagaare   & 1,885 & n/a & 46.8 & 69.8 & 54.8 & 37.2 & 34.9 & 34.2 & 34.4 & \textbf{28.9} \\
Dagbani   & 1,838 & n/a & n/a & 78.0 & 59.3 & 42.5 & 34.1 & 34.3 & 33.1 & \textbf{29.9} \\
Ewe       & 1,891 & n/a & 50.4 & 77.5 & 49.5 & 36.0 & 32.9 & 30.8 & 32.4 & \textbf{29.1} \\
Fula      & 2,391 & n/a & 55.5 & 59.5 & 52.7 & 47.0 & 42.7 & 39.8 & \textbf{37.4} & 38.7 \\
Ikposo    & 1,845 & n/a & n/a & 91.1 & 84.3 & 76.3 & 72.9 & 70.7 & 69.2 & \textbf{69.2} \\
Lingala   & 1,695 & 78.4 & \underline{29.5} & 37.6 & 35.3 & 43.2 & 35.4 & 36.1 & 30.7 & \textbf{30.1} \\
Luganda   & 601 & n/a & 32.2 & 43.3 & 31.9 & 39.8 & 24.9 & 16.7 & 14.8 & \textbf{11.2} \\
Malagasy  & 2,214 & 100.8 & 41.0 & 42.8 & 21.8 & 24.0 & 18.8 & 17.6 & 16.4 & \textbf{11.7} \\
Masaaba   & 843 & n/a & 71.1 & 72.6 & 66.3 & 60.6 & 58.2 & 47.9 & 47.1 & \textbf{45.0} \\
Nyankole  & 848 & n/a & 53.3 & 59.9 & 42.9 & 50.6 & 46.4 & 38.1 & 35.4 & \textbf{31.2} \\
Oromo     & 3,776 & n/a & 31.9 & 51.1 & 44.6 & 29.3 & 24.6 & 25.5 & 25.5 & \textbf{20.7} \\
Shona     & 1,704 & 110.8 & 33.3 & 34.2 & 25.4 & 31.1 & 26.4 & 23.4 & 23.3 & \textbf{21.3} \\
Sidama    & 3,561 & n/a & 33.8 & 78.6 & 72.7 & 36.4 & 31.1 & 35.5 & 32.0 & \textbf{28.2} \\
Soga      & 766 & n/a & 61.4 & 69.3 & 64.1 & 64.0 & 54.3 & 46.9 & 44.0 & \textbf{42.8} \\
Tigrinya  & 5,034 & n/a & 65.6 & 70.2 & 61.3 & 51.3 & 43.4 & 44.7 & 45.9 & \textbf{37.0} \\
Wolaytta  & 3,014 & n/a & n/a & 92.7 & 85.2 & 44.7 & 40.1 & 38.2 & 40.0 & \textbf{36.6} \\
\midrule
\textbf{Macro} & 39,354 & -- & -- & 63.5 & 52.3 & 44.8 & 38.3 & 36.2 & 35.0 & \textbf{31.5} \\
\bottomrule
\end{tabular}
\end{adjustbox}
\caption{Test WER (\%) on 19 WAXAL languages. $n$: test utterances scored (1.5 to 40\,s; \S\ref{sec:setup}). n/a: language not supported by the model (no Whisper language or MMS adapter). MMS-1B is \texttt{mms-1b-all} with each language's adapter. Omni-1B is Omnilingual's 1B CTC model and Omni-7B its 7B LLM model, both without WAXAL training. $^{\ast}$Evaluated on a 500-utterance sample per language. Macro WER is given only for models covering all 19 languages. Bold: lowest WER among fine-tuned models; underlined: a zero-shot result lower than every fine-tuned model. SeamlessM4T-v2 is in Appendix~\ref{app:frontier}; CER is in Appendix~\ref{app:cer}.}
\label{tab:bench}
\end{table*}

\emph{Fine-tuning is needed for usable accuracy on most languages, though zero-shot models are already competitive on a few.} Table~\ref{tab:bench} reports WER for all models. Zero-shot performance varies widely. Whisper-Large-v3 scores between 78 and 117 WER on the four languages it supports, and SeamlessM4T-v2 supports only five (Appendix~\ref{app:frontier}). MMS-1B with its language adapters and Omnilingual-7B do much better: on the 16 languages both cover, they reach the same macro WER (47.8), with MMS-1B lower on 6 of them. Omnilingual-7B, the best zero-shot model covering all 19 languages, reaches 52.3. As expected, fine-tuning on WAXAL improves substantially on these baselines: the three edge models reach 36.2 to 44.8, and Omnilingual with an LLM decoder reaches 31.5, the lowest WER among the fine-tuned models on 18 of 19 languages.

The gap between the best zero-shot and best fine-tuned model varies widely across languages, from 48.5 points on Wolaytta to none on Lingala, where MMS-1B without WAXAL training (29.5) is slightly better than every fine-tuned model (30.1); zero-shot models are also within 6 points on Shona and Sidama. This likely reflects uneven representation of these languages in pretraining data. Among the edge models, MMS-300M has lower macro WER than Whisper-Small (36.2 vs.\ 38.3; difference 2.1, 95\% interval [1.7, 2.5]). MMS-300M is significantly better on 10 languages and Whisper-Small on 4, and the remaining 5 differences are not significant.

\section{Decoding Architecture}
\label{sec:arch}

\emph{Q1: With encoder pretraining held fixed, the LLM decoder is more accurate in every language family, so the family pattern between MMS and Whisper is better explained by pretraining than by the decoder.}

\paragraph{Across models: a family pattern.} Among the edge models, the better choice between MMS-300M (CTC) and Whisper-Small (autoregressive) appears to follow language family (Table~\ref{tab:arch}). Whisper-Small is significantly better on four of the five Afro-Asiatic languages (Amharic, Tigrinya, Oromo, Sidama), while MMS-300M is significantly better on five of the six Bantu languages (the sixth, Lingala, is not significantly different) and is better or tied elsewhere. MMS-300M also has lower CER on 18 of 19 languages. However, these two models differ in pretraining data, size, and fine-tuning strategy, so the comparison cannot attribute the pattern to decoding.

\paragraph{Shared encoder: the pattern does not hold.} Omnilingual lets us vary the decoder with encoder pretraining held fixed. After fine-tuning, the LLM decoder has lower WER than the CTC decoder on 18 of 19 languages (macro 31.5 vs.\ 35.0; difference 3.5, 95\% interval [3.4, 3.7]), including all six Bantu languages. The difference is significant on 16 languages; on Lingala and Ikposo it is not, and on Fula the CTC decoder is significantly better (Figure~\ref{fig:forest}a). As Table~\ref{tab:arch} shows, the LLM decoder's advantage is present in every family. This suggests that the family pattern seen between MMS and Whisper is better explained by differences in pretraining and model setup than by CTC versus autoregressive decoding as such.

This comparison still does not isolate architecture completely. The LLM decoder adds a 1.3B-parameter Transformer decoder and decodes autoregressively. On a single CPU thread, the fine-tuned LLM model runs about four to five times slower than real time and needs 10.5\,GB of memory, while the CTC model runs about four times faster than real time in 3.2\,GB (Appendix~\ref{app:device}). Its accuracy gains therefore come with much higher memory and latency costs.

\begin{table}[t]
\centering\small
\setlength{\tabcolsep}{4pt}
\begin{adjustbox}{max width=\columnwidth}
\begin{tabular}{lccccc}
\toprule
& & \multicolumn{2}{c}{\textbf{Different pretrain.}} & \multicolumn{2}{c}{\textbf{Shared encoder}} \\
\cmidrule(lr){3-4}\cmidrule(lr){5-6}
\textbf{Family} & \# & MMS & Whisper-S & CTC & LLM \\
\midrule
Bantu         & 6 & \textbf{34.8} & 40.9 & 32.5 & \textbf{30.3} \\
Afro-Asiatic  & 5 & 35.3 & \textbf{33.8} & 35.6 & \textbf{29.5} \\
Kwa and Gur   & 5 & \textbf{40.8} & 41.6 & 40.3 & \textbf{37.3} \\
Other         & 3 & \textbf{33.0} & 35.2 & 30.3 & \textbf{27.9} \\
\bottomrule
\end{tabular}
\end{adjustbox}
\caption{Mean WER (\%) by language family for fine-tuned models. Between MMS-300M and Whisper-Small, the better model depends on family; between Omnilingual's CTC and LLM decoders, which share one encoder, it does not. Other: Acholi, Fula, Malagasy.}
\label{tab:arch}
\end{table}

\begin{figure}[t]
\centering
\begin{adjustbox}{max width=\columnwidth}
\begin{tikzpicture}
\begin{axis}[name=p0,  width=0.53\columnwidth, height=7.4cm, scale only axis=false,
  xmin=-10.7, xmax=2.5, ymin=0.3, ymax=19.7, ytick={1,...,19}, yticklabels={Wolaytta,Tigrinya,Soga,Sidama,Shona,Oromo,Nyankole,Masaaba,Malagasy,Luganda,Lingala,Ikposo,Fula,Ewe,Dagbani,Dagaare,Amharic,Akan,Acholi},
  tick label style={font=\scriptsize}, title={\footnotesize (a) LLM $-$ CTC}, title style={yshift=-4pt},
  xlabel={\scriptsize WER difference (points)}, xlabel style={yshift=3pt}, ytick style={draw=none},
  grid=major, grid style={gray!15}, axis line style={gray!60}]
\addplot[dashed, gray] coordinates {(0,0.3) (0,19.7)};
\addplot[only marks, mark=*, mark size=1.5pt, black, error bars/.cd, x dir=both, x explicit] coordinates {(-3.91,19) += (0.87,0) -= (0.89,0) (-3.26,18) += (0.45,0) -= (0.45,0) (-9.79,17) += (0.35,0) -= (0.35,0) (-5.45,16) += (0.73,0) -= (0.76,0) (-3.17,15) += (0.60,0) -= (0.52,0) (-3.28,14) += (0.41,0) -= (0.40,0) (1.36,13) += (0.52,0) -= (0.49,0) (-3.55,10) += (0.62,0) -= (0.61,0) (-4.65,9) += (0.30,0) -= (0.32,0) (-2.19,8) += (0.89,0) -= (0.87,0) (-4.15,7) += (0.66,0) -= (0.64,0) (-4.77,6) += (0.28,0) -= (0.27,0) (-1.95,5) += (0.47,0) -= (0.46,0) (-3.82,4) += (0.41,0) -= (0.41,0) (-1.12,3) += (0.98,0) -= (0.93,0) (-8.99,2) += (0.35,0) -= (0.34,0) (-3.37,1) += (0.45,0) -= (0.42,0)};
\addplot[only marks, mark=o, mark size=1.5pt, gray, error bars/.cd, x dir=both, x explicit] coordinates {(-0.05,12) += (1.20,0) -= (1.14,0) (-0.60,11) += (0.71,0) -= (0.70,0)};
\end{axis}
\begin{axis}[name=p1, at={($(p0.east)+(0.45cm,0)$)}, anchor=west, width=0.53\columnwidth, height=7.4cm, scale only axis=false,
  xmin=-3.8, xmax=4.9, ymin=0.3, ymax=19.7, ytick={1,...,19}, yticklabels={},
  tick label style={font=\scriptsize}, title={\footnotesize (b) All-19 $-$ per-lang.}, title style={yshift=-4pt},
  xlabel={\scriptsize WER difference (points)}, xlabel style={yshift=3pt}, ytick style={draw=none},
  grid=major, grid style={gray!15}, axis line style={gray!60}]
\addplot[dashed, gray] coordinates {(0,0.3) (0,19.7)};
\addplot[only marks, mark=*, mark size=1.5pt, black, error bars/.cd, x dir=both, x explicit] coordinates {(1.80,17) += (0.28,0) -= (0.28,0) (3.17,16) += (0.78,0) -= (0.76,0) (2.52,15) += (0.60,0) -= (0.63,0) (0.70,14) += (0.56,0) -= (0.44,0) (-2.69,13) += (0.49,0) -= (0.49,0) (3.06,12) += (1.24,0) -= (1.25,0) (-0.70,11) += (0.56,0) -= (0.55,0) (0.69,10) += (0.52,0) -= (0.52,0) (1.39,9) += (0.26,0) -= (0.25,0) (1.91,6) += (0.25,0) -= (0.26,0) (1.42,4) += (0.43,0) -= (0.43,0) (3.54,2) += (0.27,0) -= (0.29,0) (3.17,1) += (0.48,0) -= (0.48,0)};
\addplot[only marks, mark=o, mark size=1.5pt, gray, error bars/.cd, x dir=both, x explicit] coordinates {(-0.03,19) += (0.78,0) -= (0.82,0) (-0.15,18) += (0.41,0) -= (0.39,0) (-0.70,8) += (0.80,0) -= (0.79,0) (0.16,7) += (0.59,0) -= (0.61,0) (0.03,5) += (0.48,0) -= (0.45,0) (-0.56,3) += (0.90,0) -= (0.94,0)};
\end{axis}
\end{tikzpicture}
\end{adjustbox}
\caption{Per-language WER differences with 95\% paired bootstrap intervals. (a) Omnilingual LLM minus CTC decoder (negative: LLM better). (b) The all-19 model minus per-language models at the same 5k-step budget (positive: pooling worse). Filled: significant; hollow: interval includes zero.}
\label{fig:forest}
\end{figure}

\section{Training Granularity}
\label{sec:gran}

\emph{Q2: One model for all 19 languages comes within about one WER point of nineteen per-language models, a modest accuracy cost for a large saving in storage and maintenance.} Can one model replace many? Using Omnilingual with the LLM decoder and the same 5,000-step budget, we train one model per language family for the five families with more than one WAXAL language, and one model on all 19 languages. Each is evaluated on the test set of every language it was trained on (Table~\ref{tab:gran}; per-language results in Appendix~\ref{app:gran}).

\begin{table}[t]
\centering\small
\setlength{\tabcolsep}{3.5pt}
\begin{adjustbox}{max width=\columnwidth}
\begin{tabular}{lcccccc}
\toprule
\textbf{Pooled model} & \# & \textbf{\shortstack{Pooled\\steps}} & \textbf{Per-lang.} & \textbf{Pooled} & $\Delta$ & \textbf{W/B} \\
\midrule
Bantu       & 6 & 5k & 30.3 & 31.1 & +0.8 & 4/0 \\
Kwa         & 3 & 5k & 42.5 & 43.7 & +1.2 & 3/0 \\
Gur         & 2 & 5k & 29.4 & 29.8 & +0.4 & 1/0 \\
Cushitic    & 2 & 5k & 24.4 & 24.3 & $-$0.1 & 0/0 \\
Semitic     & 2 & 5k & 31.0 & 30.7 & $-$0.3 & 0/2 \\
\midrule
All 19      & 19 & 5k & \multirow{2}{*}{31.5} & 32.5 & +1.0 [+0.8, +1.1] & 11/2 \\
All 19      & 19 & 20k &  & 32.2 & +0.7 [+0.5, +0.8] & 10/2 \\
\bottomrule
\end{tabular}
\end{adjustbox}
\caption{\emph{Take-home: one model for all 19 languages comes within about one WER point of nineteen per-language models, a small cost for serving every language with a single model; training it four times longer narrows the gap only slightly.} Macro test WER (\%) of per-language versus pooled models (Omnilingual LLM) over the languages each pooled model covers. Per-lang.\ is always the per-language models trained for 5k steps each (19 models, 95k steps in total); no per-language model was trained for 20k steps, so both all-19 rows share one baseline. Pooled steps: training steps of the single pooled model. $\Delta$: pooled minus per-language (positive: pooled is worse), with a 95\% paired bootstrap interval for the all-19 models; over the 15 languages with a family model, $\Delta = +0.6$ $[+0.4, +0.7]$. W/B: languages where the pooled model is significantly worse / better.}
\label{tab:gran}
\end{table}

Family pooling stays within about one point of per-language training. It is slightly worse for the Bantu, Kwa, and Gur families, significantly so on 8 of their 11 languages, and unchanged or slightly better for the two Afro-Asiatic families, where it is significantly better on both Semitic languages. Over the 15 languages with a family model, pooling costs 0.6 WER on average (95\% interval [0.4, 0.7]). The all-19 model is 1.0 points worse on average, significantly worse on 11 languages and better on 2 (Figure~\ref{fig:forest}b). At a fixed step count the pooled model sees each language far less often (about one pass over the pooled data, versus 10 to 70 passes for per-language models), so we also trained it for four times as many steps. This narrowed the gap only to 0.7, which suggests the difference is not simply undertraining. Even at 20k steps, the pooled model uses about a fifth of the total compute of the nineteen per-language models (19 $\times$ 5k = 95k steps). With more steps, several higher-resource languages improved (Tigrinya from 40.5 to 36.7) while some lower-resource ones got worse (Acholi from 33.1 to 35.0, Masaaba from 44.3 to 46.6). These results are consistent with earlier findings for multilingual ASR~\citep{pratap2020massively}. In practice, one model covering all 19 languages performs close to nineteen dedicated models at a fraction of the storage and maintenance cost, but its small loss is consistent across languages, and on average we found no setting in which pooling improved on per-language training.

\section{Generalization}
\label{sec:gen}

\paragraph{Domain shift.} \emph{Q3: Fine-tuned models remain usable on read speech, although the pretrained models do better there.} We evaluate on FLEURS read speech for the six WAXAL languages it covers (Table~\ref{tab:fleurs}). Models fine-tuned on WAXAL remain usable, with the best fine-tuned model per language at 22 to 65 WER, and they rank in the same order as on WAXAL. The pretrained models do better on read speech, however. Omnilingual-7B has the lowest WER on five of the six languages and MMS-1B on the sixth (Luganda), and the 300M Omnilingual LLM model before WAXAL fine-tuning (36.7 macro) is better than the same model after it (42.0) on five of the six languages. This is consistent with read speech being closer to the pretraining data of these models, which may include FLEURS itself, and suggests that fine-tuning on spontaneous speech trades some read-speech accuracy for in-domain gains. Whisper-Large-v3 has WER of 73 or more on the three FLEURS languages it supports (Appendix~\ref{app:frontier}).

\begin{table}[t]
\centering\small
\setlength{\tabcolsep}{2.6pt}
\begin{adjustbox}{max width=\columnwidth}
\begin{tabular}{lcccc|ccccc}
\toprule
& \multicolumn{4}{c|}{\textbf{Pretrained}} & \multicolumn{5}{c}{\textbf{Fine-tuned on WAXAL}} \\
\textbf{Lang.} & MMS-1B & Omni-1B & Omni-LLM & Omni-7B & Wh-T & Wh-S & MMS & O-CTC & O-LLM \\
\midrule
Amharic   & 27.7 & 37.9 & 23.1 & \textbf{15.9} & 41.2 & 29.8 & 31.2 & 29.1 & 22.3 \\
Fula      & 47.7 & 52.0 & 47.2 & \textbf{43.6} & 77.1 & 67.1 & 56.6 & 53.0 & 51.8 \\
Lingala   & 15.1 & 17.0 & 16.5 & \textbf{12.1} & 76.1 & 50.0 & 38.2 & 23.1 & 23.4 \\
Luganda   & \textbf{41.1} & 44.6 & 43.7 & 41.6 & 70.8 & 64.0 & 55.8 & 53.1 & 51.3 \\
Oromo     & 65.9 & 71.5 & 64.9 & \textbf{59.8} & 76.4 & 65.4 & 69.1 & 71.5 & 71.2 \\
Shona     & 23.5 & 23.6 & 25.0 & \textbf{15.2} & 73.5 & 57.8 & 40.9 & 36.2 & 31.9 \\
\midrule
\textbf{Macro} & 36.8 & 41.1 & 36.7 & 31.4 & 69.2 & 55.7 & 48.6 & 44.3 & 42.0 \\
\bottomrule
\end{tabular}
\end{adjustbox}
\caption{WER (\%) on FLEURS read speech, 300 utterances per language except Oromo (41). MMS-1B: with each language's adapter; Omni-1B: CTC 1B; Omni-LLM: LLM 300M before WAXAL fine-tuning; O-CTC and O-LLM: the same 300M models after per-language WAXAL fine-tuning. Bold: lowest WER per language.}
\label{tab:fleurs}
\end{table}

\paragraph{Language shift.} \emph{Q4: Neither training on other languages nor in-context examples replaced in-language data when the model was used directly.} Can data from other languages help a language with none? For seven languages, chosen before running the experiment (the highest-resource member of each multi-language family, and all four languages without a WAXAL relative), we fine-tune Omnilingual on the other 18 languages and evaluate on the held-out one (Table~\ref{tab:loo}; all Omnilingual columns are scored on the same 500-utterance sample). In every case the held-out language is transcribed worse than by the original pretrained model, by 6 to 47 points. Because Omnilingual was pretrained on these languages, this is consistent with catastrophic forgetting~\citep{kirkpatrick2017overcoming}.

Omnilingual also offers a 7B model that adapts to a language from a few transcribed examples given in context~\citep{omnilingual2025}. Given ten training utterances (5 to 20\,s) of each held-out language, it has lower WER than both the pretrained 300M model and the leave-one-out models on all seven languages. It still remains 6 to 37 points behind per-language fine-tuning (19.4 points on average, 95\% interval [18.7, 20.0]), with a model more than four times larger.

To test languages absent from pretraining, we repeat the leave-one-out experiment with Whisper-Small for three languages it does not support. WER stays between 84 and 132. With the same recipe applied to the held-out language's own data, WER is 78.5 for Ewe and 55.1 for Fula, and the released per-language models, trained with a different recipe, reach 33 and 43. For Luganda the matched comparison is not available: Luganda is not a Whisper language, the recipe decodes without a language token, and with 5.4k training utterances the monolingual model often fell back to English output (124 of 638 test utterances), which we do not treat as a meaningful WER. Luganda, which has Bantu relatives in the training set, fares best among the held-out Whisper runs, but with three languages we draw no conclusion about relatedness. In our experiments, neither related-language data nor in-context examples substituted for in-language fine-tuning when the resulting model was used directly, and a model fine-tuned on some languages should not be assumed to keep the pretrained model's ability on others.

\begin{table}[t]
\centering\small
\setlength{\tabcolsep}{3.5pt}
\begin{adjustbox}{max width=\columnwidth}
\begin{tabular}{lrcccc}
\toprule
\textbf{Held-out} & $n$ & \textbf{Pretr.} & \textbf{Other 18} & \textbf{7B + 10 ex.} & \textbf{Own data} \\
\midrule
\multicolumn{6}{l}{\emph{Omnilingual (language in pretraining)}} \\
Lingala   & 485 & 43.7 & 59.6 & 36.0 & 29.7 \\
Ewe       & 500 & 57.2 & 96.7 & 53.7 & 28.7 \\
Tigrinya  & 500 & 74.3 & 92.0 & 64.0 & 37.7 \\
Fula      & 500 & 58.4 & 87.1 & 51.2 & 38.6 \\
Acholi    & 479 & 64.8 & 90.8 & 46.4 & 33.0 \\
Malagasy  & 500 & 28.5 & 75.5 & 26.9 & 11.7 \\
Wolaytta  & 500 & 81.3 & 87.2 & 72.4 & 35.8 \\
\midrule
\multicolumn{6}{l}{\emph{Whisper-Small (language not supported)}} \\
Luganda   & 601 & n/a & 84.1 & -- & n/a$^\ddagger$ (24.9) \\
Ewe       & 1,891 & n/a & 102.0 & -- & 78.5 (32.9) \\
Fula      & 2,391 & n/a & 132.0 & -- & 55.1 (42.7) \\
\bottomrule
\end{tabular}
\end{adjustbox}
\caption{WER (\%) on held-out languages. Pretr.: Omnilingual LLM-300M without WAXAL training. Other 18: fine-tuned on the remaining 18 languages. 7B + 10 ex.: Omnilingual's in-context 7B model given ten transcribed training utterances of the language. Own data: fine-tuned on the language itself. For Omnilingual, all four columns are scored on the same $n$ utterances (the 500-utterance sample) and Own data is the per-language model. For Whisper, Own data uses the same recipe as Other 18, with the released per-language model, trained with a different recipe, in parentheses, on the full test set. $^\ddagger$Not reported: without a language token, this model often decoded Luganda audio as English (\S\ref{sec:gen}).}
\label{tab:loo}
\end{table}

\section{Native-Speaker Audit}
\label{sec:audit}

Native speakers of all 19 languages (one to four per language, volunteering through university and community networks) reviewed model outputs. For each language and edge model, they examined 40 test utterances, 20 from the best-performing quartile and 20 from the worst, comparing audio, reference, and hypothesis and recording error types with a shared template (Appendix~\ref{app:audit}). Annotators were native speakers rather than trained linguists, and multi-annotator languages were reviewed jointly, so we treat these findings as exploratory.

\paragraph{Failure modes.} Whisper models were decoded greedily without a repetition penalty, so that their natural behavior was visible. Annotators found unbounded repetition loops in 14 of 19 languages, more often in Whisper-Tiny; in a Soga example the model transcribes the start correctly and then repeats ``emiti'' over 100 times. Measured automatically over the full test sets (a sequence of one to four words repeated at least four times in a row that does not occur in the reference), loops affect 1.2\% of utterances for Whisper-Tiny and 0.6\% for Whisper-Small, against 0.01\% for the Omnilingual LLM decoder and almost none for the CTC models. MMS-300M, as a CTC model, emits at most one token per audio frame and cannot produce unbounded loops. Its typical errors are phonetic approximations such as compound splitting (Acholi ``kacato'' $\rightarrow$ ``ka cato'') and fricative confusion (``ovarol'' $\rightarrow$ ``ofarol''), which a reader can often recover from.

\paragraph{Script.} CER/WER ratios are higher for the Ge'ez-script languages than for Latin-script Bantu languages: for the Omnilingual LLM model, 0.41 for Amharic and 0.43 for Tigrinya, against 0.17 for Shona and 0.25 for Luganda (MMS-300M: 0.37 and 0.41, against 0.17 and 0.20). Each Ge'ez character encodes a consonant--vowel syllable, so words contain few characters and a single wrong vowel changes both a character and the word. Error rates are therefore not directly comparable across scripts, and, following \citet{k-etal-2025-advocating}, we report CER alongside WER (Appendix~\ref{app:cer}).

\paragraph{Orthography.} Annotators suggested that special letters and diacritics, which Kwa and Gur orthographies use to mark vowel quality and tone (e.g.\ \textipa{E}, \textipa{O}, \textipa{N}), contribute to errors. Measured across the 17 Latin-script languages, however, the share of such characters in the references does not predict error rates (Spearman $\rho = -0.04$ for WER and $0.28$ for CER of the Omnilingual LLM model, both $p > 0.25$), and without Ikposo the Kwa and Gur languages have WER similar to the Bantu languages (29.3 vs.\ 30.3). Ikposo is the hardest language for every model (69 to 76 WER after fine-tuning). Its references write one vowel with two visually near-identical letters, \textipa{U} (U+028A) and \textscriptv{} (U+028B), but merging them lowers WER by less than one point. Merging these letters with \emph{u} and \textiota{} with \emph{i} lowers WER by 10 to 13 points for every fine-tuned model, so vowel-quality distinctions account for part of the difficulty, but WER stays near 60.

\section{Discussion}
\label{sec:disc}

For spontaneous speech in most of these languages, in-language fine-tuning was necessary for usable accuracy. Zero-shot models are competitive on a few languages (MMS-1B reaches 29.5 on Lingala; Omnilingual-7B reaches 21.8 on Malagasy and 25.4 on Shona), and in-context examples help the 7B model on held-out languages, but on average they remain well behind fine-tuned models a fraction of their size, and the 7B model is too large for most local deployment. Beyond that, the choices we studied involve trade-offs rather than clear winners. The LLM decoder is the most accurate but needs 1.6B parameters and runs slower than real time on a single CPU thread. A single multilingual model is slightly but consistently worse than per-language models and much cheaper to maintain. CTC models avoid repetition loops and their errors are easier to recover from.

\begin{figure}[t]
\centering
\begin{tikzpicture}
\begin{axis}[width=\columnwidth, height=5.2cm, xmode=log, xmin=0.03, xmax=12, ymin=28, ymax=48,
  xlabel={Real-time factor, single CPU thread (log scale)}, ylabel={Macro WER (\%)},
  xtick={0.1,1,10}, xticklabels={0.1,1,10}, tick label style={font=\small}, label style={font=\small},
  grid=major, grid style={gray!20}]
\addplot[dashed, gray] coordinates {(1,28) (1,48)};
\node[font=\scriptsize, gray, anchor=north west] at (axis cs:1.05,47.6) {real time};
\addplot[only marks, mark=*, mark size=2.2pt, color=black] coordinates {(0.0568,44.81)} node[anchor=west, inner sep=3pt, font=\scriptsize, align=center] {Whisper-Tiny\\1.0\,GB};
\addplot[only marks, mark=*, mark size=2.2pt, color=black] coordinates {(0.3286,38.32)} node[anchor=south west, inner sep=3pt, font=\scriptsize, align=center] {Whisper-Small\\1.9\,GB};
\addplot[only marks, mark=*, mark size=2.2pt, color=black] coordinates {(0.2646,36.25)} node[anchor=east, inner sep=3pt, font=\scriptsize, align=center] {MMS-300M\\2.8\,GB};
\addplot[only marks, mark=*, mark size=2.2pt, color=black] coordinates {(0.2718,35.04)} node[anchor=north west, inner sep=3pt, font=\scriptsize, align=center] {Omni-CTC\\3.2\,GB};
\addplot[only marks, mark=*, mark size=2.2pt, color=black] coordinates {(4.7683,31.53)} node[anchor=south east, inner sep=3pt, font=\scriptsize, align=center] {Omni-LLM\\10.5\,GB};
\end{axis}
\end{tikzpicture}
\caption{Accuracy and CPU cost of the fine-tuned models: macro test WER against single-thread real-time factor on 30\,s of speech (left of the dashed line is faster than real time), labelled with peak memory. Data in Table~\ref{tab:bench} and Appendix~\ref{app:device}.}
\label{fig:cost}
\end{figure}

For deployment on constrained hardware (Figure~\ref{fig:cost}), the Whisper and CTC models run faster than real time on a single CPU thread in 1.0 to 3.2\,GB of memory (Appendix~\ref{app:device}). Whisper-Tiny suits the tightest budgets, while MMS-300M and the Omnilingual CTC model offer better accuracy without repetition loops at similar cost. The Omnilingual LLM models are the most accurate but would need compression or a GPU for comparable use. Finally, where in-language data is available, it is worth using for every language a model is expected to serve: in our experiments, models trained on other languages, even related ones, were far less accurate when applied directly to a held-out language. We did not test whether such a multilingual model makes a better starting point for fine-tuning on a small amount of in-language data, which may matter more in practice, where data is scarce.

\section*{Limitations}
Each condition was trained once. Our bootstrap intervals capture variation over test utterances but not over training runs, so differences of about one WER point, including several in Section~\ref{sec:gran}, may not hold across training seeds. The Omnilingual-7B, SeamlessM4T-v2, and Omnilingual held-out results use 500-utterance samples. Our pooling and held-out results should not be read as general statements about cross-lingual transfer: Omnilingual was already pretrained on all 19 languages, so transfer may have occurred during pretraining; each language has 25 to 190 hours of training data, whereas multilingual training is known to help most when data is scarcer; and our language families are broad groupings. The held-out experiments cover seven languages for Omnilingual and three for Whisper, and the Whisper comparison uses different training recipes for the held-out and per-language models. The edge models and Omnilingual models use different training procedures, so comparisons between them reflect full pipelines rather than architecture alone. The native-speaker audit covers the edge models only, uses 40 utterances per language and model, and reports no inter-annotator agreement. Several languages have few unique test speakers. WER also penalizes hesitations and code-switching that may not affect intelligibility. Error rates remain too high for high-stakes uses such as healthcare without further validation.

\section*{Ethical Considerations}
WAXAL was collected with informed consent and released under CC-BY-4.0~\citep{Diacketal2026}. Native-speaker contributors volunteered to help build recognition for their own languages and agreed to their judgments being used in this work. They were not paid; contributors who met a contribution threshold are offered co-authorship and the others are acknowledged. The audit involved only reviewing model outputs on an existing, consented corpus and collected no personal data from annotators. Released models are licensed under CC-BY-4.0, except models fine-tuned from MMS, which keep the base model's CC-BY-NC-4.0 license, and include usage guidelines that prohibit non-consensual applications such as voice cloning or surveillance. In line with the ACL policy on AI writing assistance, we disclose that AI assistants were used for code, LaTeX formatting, and editing; all experimental design, analysis, and interpretation were carried out and checked by the authors.

\section*{Acknowledgements}
This work was supported by LyngualLabs (compute, researchers, and storage), Open Token (compute), and CMU Africa (researchers and native speakers). 
We thank the native-speaker contributors who volunteered their time to review model outputs in their languages.

\bibliography{reference}

\appendix
\raggedbottom

\section{Dataset Statistics}
\label{app:dataset}
Table~\ref{tab:app-dataset} lists the 19 languages with their families, scripts, and split sizes after cleaning.

\begin{table*}[h]
\centering\small
\begin{adjustbox}{max width=\textwidth}
\begin{tabular}{llllrrrrr}
\toprule
\textbf{Language} & \textbf{ISO} & \textbf{Family} & \textbf{Script} & \textbf{Train utt.} & \textbf{Train h} & \textbf{Dev utt.} & \textbf{Test utt.} & \textbf{Scored} \\
\midrule
Acholi    & ach & Nilotic & Latin & 4,108 & 25.8 & 519 & 528 & 506 \\
Akan      & aka & Kwa & Latin & 10,107 & 55.0 & 1,123 & 1,522 & 1,522 \\
Amharic   & amh & Semitic & Ge'ez & 38,022 & 189.7 & 2,912 & 3,420 & 3,420 \\
Dagaare   & dga & Gur & Latin & 15,071 & 83.4 & 1,893 & 1,910 & 1,885 \\
Dagbani   & dag & Gur & Latin & 14,231 & 77.0 & 1,750 & 1,838 & 1,838 \\
Ewe       & ewe & Kwa & Latin & 15,052 & 79.5 & 1,916 & 1,891 & 1,891 \\
Fula      & ful & Atlantic & Latin & 19,132 & 99.7 & 2,358 & 2,391 & 2,391 \\
Ikposo    & kpo & Kwa & Latin & 14,415 & 82.1 & 1,760 & 1,845 & 1,845 \\
Lingala   & lin & Bantu & Latin & 14,399 & 71.9 & 1,844 & 1,866 & 1,695 \\
Luganda   & lug & Bantu & Latin & 5,455 & 37.3 & 664 & 638 & 601 \\
Malagasy  & mlg & Austronesian & Latin & 18,526 & 94.7 & 2,270 & 2,290 & 2,214 \\
Masaaba   & mas & Bantu & Latin & 6,865 & 39.3 & 849 & 858 & 843 \\
Nyankole  & nyn & Bantu & Latin & 6,783 & 40.8 & 831 & 858 & 848 \\
Oromo     & orm & Cushitic & Latin & 38,182 & 189.9 & 3,078 & 3,777 & 3,776 \\
Shona     & sna & Bantu & Latin & 14,109 & 79.7 & 1,727 & 1,749 & 1,704 \\
Sidama    & sid & Cushitic & Latin & 38,752 & 192.7 & 3,091 & 3,561 & 3,561 \\
Soga      & sog & Bantu & Latin & 6,219 & 40.0 & 841 & 817 & 766 \\
Tigrinya  & tir & Semitic & Ge'ez & 40,856 & 181.9 & 4,425 & 5,034 & 5,034 \\
Wolaytta  & wal & Omotic & Latin & 41,833 & 197.3 & 2,444 & 3,014 & 3,014 \\
\midrule
\textbf{Total} & & & & 362,117 & 1,858 & 36,295 & 39,807 & 39,354 \\
\bottomrule
\end{tabular}
\end{adjustbox}
\caption{WAXAL data used in this work, after removing utterances whose transcript would need more than 4 words per second (\S\ref{sec:setup}). Scored: test utterances of 1.5 to 40\,s on which every model was evaluated.}
\label{tab:app-dataset}
\end{table*}

\section{Training Details}
\label{app:training}

\paragraph{Edge models.} Whisper-Tiny, Whisper-Small, and MMS-300M are trained per language with the settings in \S\ref{sec:setup}, with MMS-300M's pretrained encoder kept frozen: effective batch size 32 via gradient accumulation, learning rate $10^{-4}$ with 500 linear warmup steps and polynomial decay, and up to 30 epochs with early stopping (patience 3), on NVIDIA A100 GPUs. Table~\ref{tab:app-convergence} gives the end-of-training validation results.

\begin{table*}[h]
\centering\small
\begin{adjustbox}{max width=\textwidth}
\begin{tabular}{l|ccc|ccc|ccc}
\toprule
& \multicolumn{3}{c|}{\textbf{MMS-300M}} & \multicolumn{3}{c|}{\textbf{Whisper-Small}} & \multicolumn{3}{c}{\textbf{Whisper-Tiny}} \\
\textbf{Language} & \textbf{Dev WER} & \textbf{Dev loss} & \textbf{Steps} & \textbf{Dev WER} & \textbf{Dev loss} & \textbf{Steps} & \textbf{Dev WER} & \textbf{Dev loss} & \textbf{Steps} \\
\midrule
Acholi   & 40.7 & 0.612 & 1000 & 40.1 & 0.801 & 500  & 53.3 & 0.890 & 500  \\
Akan     & 30.4 & 0.371 & 2000 & 30.1 & 0.419 & 1000 & 35.4 & 0.476 & 1000 \\
Amharic  & 30.9 & 0.427 & 6000 & 49.1 & 0.100 & 4500 & 54.9 & 0.130 & 5500 \\
Dagaare  & 34.2 & 0.512 & 3000 & 33.5 & 0.457 & 1500 & 36.5 & 0.516 & 1500 \\
Dagbani  & 35.0 & 0.411 & 3000 & 34.7 & 0.401 & 1000 & 41.0 & 0.474 & 1500 \\
Ewe      & 31.6 & 0.415 & 3000 & 33.5 & 0.414 & 1000 & 37.9 & 0.459 & 1500 \\
Fula     & 40.0 & 0.366 & 3500 & 42.0 & 0.531 & 1500 & 45.7 & 0.597 & 1500 \\
Ikposo   & 70.9 & 0.909 & 4000 & 73.9 & 0.485 & 1000 & 75.4 & 0.559 & 1500 \\
Lingala  & 37.2 & 0.589 & 3500 & 34.8 & 0.568 & 1000 & 40.7 & 0.646 & 1000 \\
Luganda  & 16.6 & 0.147 & 2000 & 22.2 & 0.324 & 500  & 29.4 & 0.427 & 1000 \\
Malagasy & 13.6 & 0.121 & 3000 & 19.6 & 0.308 & 1500 & 29.0 & 0.354 & 1500 \\
Masaaba  & 49.9 & 0.447 & 2500 & 59.6 & 0.816 & 500  & 60.6 & 0.927 & 1000 \\
Nyankole & 37.3 & 0.378 & 1500 & 43.3 & 0.557 & 500  & 43.8 & 0.651 & 1000 \\
Oromo    & 30.1 & 0.321 & 3500 & 28.8 & 0.418 & 3000 & 34.0 & 0.480 & 3500 \\
Shona    & 24.4 & 0.204 & 2500 & 30.9 & 0.389 & 1000 & 36.5 & 0.440 & 1500 \\
Sidama   & 40.6 & 0.295 & 4500 & 38.1 & 0.631 & 3500 & 42.9 & 0.704 & 3000 \\
Soga     & 45.5 & 0.320 & 2500 & 50.9 & 0.605 & 500  & 58.8 & 0.684 & 500  \\
Tigrinya & 43.1 & 1.194 & 5500 & 50.7 & 0.172 & 3500 & 57.5 & 0.217 & 6000 \\
Wolaytta & 32.4 & 0.240 & 4500 & 34.3 & 0.452 & 3000 & 36.9 & 0.495 & 3000 \\
\bottomrule
\end{tabular}
\end{adjustbox}
\caption{Edge-model training: WER (\%) and loss on the WAXAL validation split at the end of training, and total training steps, as logged during training. Steps vary with dataset size and early stopping. Validation WER uses the training-time normalization and is not comparable to the test results in Table~\ref{tab:bench}.}
\label{tab:app-convergence}
\end{table*}

\paragraph{Omnilingual.} We use the official fine-tuning recipes in \texttt{fairseq2} with the \texttt{omniASR\_tokenizer\_v1} tokenizer: AdamW with learning rate $10^{-5}$ and dynamic batches of up to 15M audio samples with gradient accumulation of 2 (CTC), or learning rate $5\times10^{-5}$, weight decay 0.01, and batches of up to 3M samples with accumulation of 4 (LLM); bfloat16, audio up to 30\,s, on two H200 GPUs, with the encoder trained from the first step. Every per-language, family, and all-19 model is evaluated at its final checkpoint (5,000 steps, or 20,000 for the longer all-19 run). The leave-one-out models use the same recipe on the 18 remaining languages. For in-context adaptation we use \texttt{omniASR\_LLM\_7B\_ZS} with ten examples drawn with a fixed seed from the training split, restricted to 5 to 20\,s since typical WAXAL utterances are 16 to 21\,s long.

\paragraph{Whisper held-out runs.} The leave-one-out Whisper-Small models are trained on the 18 remaining languages, capped at 3,000 utterances per language, for 4,000 steps (batch 16, learning rate $10^{-5}$, 200 warmup steps, fp16), without a language token, with labels truncated to Whisper's 448-token limit. The matched own-data runs use the same recipe on all of the language's training data. For Luganda, 124 of 638 test outputs of the own-data model were runaway English text more than three times the reference length (283 WER). Retraining for 1,400 steps, which matches the number of passes over the data used for Ewe and Fula, made this worse (192 of 638), so the failure is not overfitting; we attribute it to language identification falling back to English, which the multilingual leave-one-out model, trained on about 54k utterances without a language token, does not show.

\section{Pooled Model Results by Language}
\label{app:gran}
\begin{table}[H]
\centering\small
\begin{adjustbox}{max width=\columnwidth}
\begin{tabular}{lcccc}
\toprule
\textbf{Language} & \textbf{Per-lang.} & \textbf{Family} & \textbf{All 19 (5k)} & \textbf{All 19 (20k)} \\
\midrule
Acholi    & 33.1 & -- & 33.1 & 35.0 \\
Akan      & 29.2 & 30.0 & 29.0 & 30.1 \\
Amharic   & 25.0 & 24.7 & 26.8 & 24.4 \\
Dagaare   & 28.9 & 29.1 & 32.1 & 31.4 \\
Dagbani   & 29.9 & 30.6 & 32.4 & 31.6 \\
Ewe       & 29.1 & 30.4 & 29.8 & 29.5 \\
Fula      & 38.7 & -- & 36.1 & 36.8 \\
Ikposo    & 69.2 & 70.8 & 72.3 & 71.4 \\
Lingala   & 30.1 & 31.0 & 29.4 & 30.1 \\
Luganda   & 11.2 & 11.4 & 11.9 & 11.5 \\
Malagasy  & 11.7 & -- & 13.1 & 12.0 \\
Masaaba   & 45.0 & 46.1 & 44.3 & 46.6 \\
Nyankole  & 31.2 & 32.6 & 31.4 & 32.6 \\
Oromo     & 20.7 & 20.7 & 22.6 & 20.7 \\
Shona     & 21.3 & 21.9 & 21.3 & 21.5 \\
Sidama    & 28.2 & 28.0 & 29.6 & 28.1 \\
Soga      & 42.8 & 43.7 & 42.3 & 43.8 \\
Tigrinya  & 37.0 & 36.7 & 40.5 & 36.7 \\
Wolaytta  & 36.6 & -- & 39.8 & 37.7 \\
\bottomrule
\end{tabular}
\end{adjustbox}
\caption{Per-language test WER (\%) of per-language and pooled Omnilingual LLM models. --: no family model (no WAXAL relative).}
\label{tab:app-gran}
\end{table}

\section{Additional Baselines}
\label{app:frontier}
\paragraph{SeamlessM4T-v2} on its five supported WAXAL languages (500-utterance samples): Luganda 27.7, Shona 86.4, Amharic 99.0, Oromo 106.9, and Fula 114.2 WER.

\paragraph{Omnilingual 300M models without fine-tuning.} The CTC model reaches a macro WER of 68.2 on WAXAL. The LLM model is reported for the seven held-out languages in Table~\ref{tab:loo}.

\paragraph{Whisper-Large-v3 on FLEURS.} With the language token forced, Whisper-Large-v3 scores 164.1 (Amharic), 73.1 (Lingala), and 115.3 (Shona) WER on the FLEURS utterances of Table~\ref{tab:fleurs}; it does not support Fula, Luganda, or Oromo.

\section{Sample-Based Evaluation Check}
\label{app:sample}
To check the 500-utterance sampling used for the largest models, we compare full-test-set WER with WER on the sampled utterances alone for three models evaluated on the full test sets. Macro WER is 68.2 vs.\ 68.0 for the pretrained Omnilingual CTC-300M model, 31.5 vs.\ 31.4 for the fine-tuned Omnilingual LLM model, and 36.2 vs.\ 36.1 for MMS-300M, with mean absolute per-language differences of 0.5 to 0.6 points (at most 3.1 for any one language).

\section{Inference Cost}
\label{app:device}
Table~\ref{tab:app-device} reports single-thread CPU cost for the five fine-tuned model types on 10\,s and 30\,s segments of real Luganda test speech, each model measured in a separate process on an otherwise idle machine.

\begin{table}[H]
\centering\small
\begin{adjustbox}{max width=\columnwidth}
\begin{tabular}{lcccc}
\toprule
\textbf{Model} & \textbf{Params} & \textbf{RTF 10\,s} & \textbf{RTF 30\,s} & \textbf{RAM (GB)} \\
\midrule
Whisper-Tiny   & 38M & 0.10 & 0.06 & 1.0 \\
Whisper-Small  & 242M & 0.61 & 0.33 & 1.9 \\
MMS-300M       & 315M & 0.24 & 0.26 & 2.8 \\
Omni-CTC       & 325M & 0.24 & 0.27 & 3.2 \\
Omni-LLM       & 1,628M & 3.98 & 4.77 & 10.5 \\
\bottomrule
\end{tabular}
\end{adjustbox}
\caption{Single-thread CPU inference cost (INTEL XEON PLATINUM 8558, fp32, greedy decoding) on real Luganda speech: real-time factor (processing time divided by audio duration; below 1 is faster than real time, mean of 3 runs after warm-up) and peak process memory. Omni-LLM's parameter count includes its 1.3B-parameter decoder. Absolute speeds depend on the CPU (a server processor here); the relative costs between models are the main point.}
\label{tab:app-device}
\end{table}

\section{Full CER Results}
\label{app:cer}
\begin{table}[H]
\centering\small
\setlength{\tabcolsep}{4pt}
\begin{adjustbox}{max width=\columnwidth}
\begin{tabular}{lccccc}
\toprule
\textbf{Language} & Whisper-T & Whisper-S & MMS-300M & Omni-CTC & Omni-LLM \\
\midrule
Acholi    & 35.0 & 21.2 & 16.0 & 14.3 & 13.7 \\
Akan      & 15.6 & 12.4 & 10.9 & 10.2 & 9.8 \\
Amharic   & 17.9 & 12.5 & 12.0 & 12.4 & 10.3 \\
Dagaare   & 16.7 & 16.1 & 13.2 & 13.2 & 11.9 \\
Dagbani   & 17.1 & 12.6 & 11.2 & 10.9 & 10.5 \\
Ewe       & 12.6 & 11.0 & 9.4 & 9.7 & 9.1 \\
Fula      & 15.8 & 12.6 & 10.3 & 9.5 & 10.1 \\
Ikposo    & 27.9 & 25.7 & 23.3 & 22.7 & 23.5 \\
Lingala   & 23.5 & 16.0 & 14.0 & 12.3 & 12.6 \\
Luganda   & 16.8 & 8.7 & 3.4 & 2.9 & 2.8 \\
Malagasy  & 7.4 & 4.6 & 3.4 & 3.5 & 2.8 \\
Masaaba   & 20.8 & 18.7 & 10.5 & 10.2 & 10.0 \\
Nyankole  & 18.8 & 18.6 & 9.4 & 8.4 & 8.2 \\
Oromo     & 8.7 & 6.6 & 6.2 & 6.2 & 5.5 \\
Shona     & 8.4 & 6.0 & 4.0 & 3.9 & 3.7 \\
Sidama    & 10.2 & 8.1 & 8.2 & 7.5 & 7.0 \\
Soga      & 22.6 & 14.4 & 9.6 & 8.4 & 8.4 \\
Tigrinya  & 25.7 & 20.5 & 18.3 & 18.4 & 16.0 \\
Wolaytta  & 14.4 & 12.1 & 11.0 & 11.2 & 10.9 \\
\midrule
\textbf{Macro} & 17.7 & 13.6 & 10.7 & 10.3 & 9.8 \\
\bottomrule
\end{tabular}
\end{adjustbox}
\caption{Test CER (\%) of fine-tuned models, on the same utterances as Table~\ref{tab:bench}.}
\label{tab:app-cer}
\end{table}

\section{Native-Speaker Audit Briefing}
\label{app:audit}
To keep the distributed audit consistent, all native-speaker contributors received a standardized reporting template and a reference example. Because annotators were drawn from community networks and were not trained linguists, the written material was supplemented with a verbal calibration session, in which the research team discussed the error categories (e.g.\ hallucination, code-switching loss, orthographic errors) and how to identify them in model outputs.

The reference example given to annotators is reproduced below verbatim. It shows the expected structure, depth of analysis, and error categorization, using Yoruba (not one of the 19 evaluated languages) as the reference language.

\vspace{0.5em}
\noindent\textbf{Reference briefing example: Yoruba zero-shot evaluation}

\vspace{0.3em}
\noindent\textbf{Language:} Yoruba \\
\textbf{Assigned leads:} Full name(s) \\
\textbf{Models evaluated:} Whisper-Large-v3, MMS-1B, Omnilingual ASR

\paragraph{1. Overall performance ceiling and general observations.} Across all three models, achieving a 0\% word error rate on the WAXAL Yoruba test set is currently impossible due to the highly conversational, code-switched nature of the data. Modern spoken Yoruba frequently borrows from English and Pidgin, and none of the models seamlessly handle these transitions. Furthermore, tonal marking (which dictates meaning in Yoruba) remains a major vulnerability across the board.

\paragraph{2. Model-specific linguistic behaviors.}
\begin{itemize}
\itemsep0.1em
\item \textbf{Whisper-Large-v3:} Whisper is surprisingly good at recognizing when a speaker switches to English. However, it severely struggles with Yoruba orthography. It frequently drops diacritical marks (tones and subdots like \d{e}, \d{o}, \d{s}).
  \begin{itemize}
  \item \textit{Example:} When the speaker said ``B\'awo ni n\`{n}kan?'' (How are things?), Whisper transcribed it flatly as ``Bawo ni nkan'', stripping the high and low tones.
  \item \textit{Hallucinations (M2):} Because it tries to make logical sentences, if it does not understand a Yoruba slang term, it will hallucinate an English word that sounds phonetically similar, completely changing the sentence's meaning.
  \end{itemize}
\item \textbf{MMS-1B:} MMS-1B respects Yoruba tonal marks and subdots much better than Whisper. However, it completely fails on conversational speech and modern loan words. It seems heavily biased toward formal or religious Yoruba text (likely due to its pretraining data).
  \begin{itemize}
  \item \textit{Example:} When a speaker used the modern loan word ``k\d{o}mputa'' (computer), MMS-1B failed to transcribe it phonetically and instead output a string of unrelated, formal Yoruba syllables.
  \item \textit{Code-switching failure (C1 error):} It cannot handle English code-switching at all; when an English word is spoken, the model either deletes it or outputs gibberish.
  \end{itemize}
\item \textbf{Omnilingual ASR:} Omnilingual ASR balances better between formal and informal speech, but it suffers from severe orthographic errors (O1). It mixes standard Yoruba spelling conventions with outdated ones.
  \begin{itemize}
  \item \textit{Example:} It frequently misidentifies ``\d{s}'' (the sh-sound) as a regular ``s'', writing ``se'' instead of ``\d{s}e'' (to do). While a human reader can guess the meaning from context, this heavily penalizes the model's WER.
  \end{itemize}
\end{itemize}

\paragraph{3. Top error categories identified.}
\begin{itemize}
\itemsep0.1em
\item \textbf{P1 (tonal confusion):} The most common error across all models, for example confusing igb\'a (calabash), igba (two hundred), and \`igb\`a (time), because the models rely on phonetic consonants rather than pitch contours.
\item \textbf{C1 (code-switching loss):} WAXAL speakers naturally weave English into sentences (e.g.\ ``Mo f\d{e} l\d{o} s\'i market''). MMS deletes ``market'', while Whisper might try to spell ``market'' using Yoruba alphabet rules (``mak\d{e}ti''), which mismatches the reference text.
\end{itemize}

\end{document}